\documentclass[letterpaper]{article} 
\usepackage{aaai2026}  
\usepackage{times}  
\usepackage{helvet}  
\usepackage{courier}  
\usepackage[hyphens]{url}  
\usepackage{graphicx} 
\usepackage{natbib}  
\usepackage{caption} 
\usepackage{algorithm}
\usepackage{algorithmic}

\usepackage{newfloat}
\usepackage{listings}
\DeclareCaptionStyle{ruled}{labelfont=normalfont,labelsep=colon,strut=off} 
\floatstyle{ruled}
\newfloat{listing}{tb}{lst}{}
\floatname{listing}{Listing}
\usepackage{amsmath}        
\usepackage{pifont}
\newcommand{\xmark}{\ding{55}}
\usepackage{array}
\usepackage{booktabs} 
\usepackage{multirow}
\usepackage{amssymb}

\usepackage{cuted} 
\title{EasyText: Controllable Diffusion Transformer for Multilingual Text Rendering}
\author{
    Runnan Lu\textsuperscript{\rm 1}\, Yuxuan Zhang\textsuperscript{\rm 2}\, Jiaming Liu\textsuperscript{\rm 3}\, Haofan Wang\textsuperscript{\rm 4}\, Yiren Song\textsuperscript{\rm 1}\thanks{Corresponding Author.}\
}
\affiliations{
    \textsuperscript{\rm 1}National University of Singapore\\

    \textsuperscript{\rm 2}The Chinese University of Hong Kong\\
    \textsuperscript{\rm 3}Alibaba\\
    \textsuperscript{\rm 4}Liblib AI\\
    songyiren725@gmail.com
}

\begin{document}

\maketitle


\setlength{\tabcolsep}{2pt}
\begin{table*}[t]
  \centering
  \fontsize{9pt}{11pt}\selectfont  
  \begin{tabular}{l>{\centering\arraybackslash}p{1.9cm}>{\centering\arraybackslash}p{1.9cm}>{\centering\arraybackslash}p{1.7cm}>{\centering\arraybackslash}p{1.9cm}>{\centering\arraybackslash}p{1.9cm}>{\centering\arraybackslash}p{1.9cm}}
    \toprule
    \textbf{Functionality} & Position Control & Irregular Regions & Multi-Lingual & Long Text Rendering & Text-Image Blending & Unseen Characters\\
    \midrule
    Glyph-SDXL-v2   & \checkmark & \xmark & \checkmark & \checkmark & \xmark & \xmark\\
    AnyText         & \checkmark & \checkmark & \checkmark & \xmark & \checkmark & \xmark\\
    SD3.5           & \xmark & \xmark & \xmark & \xmark & \checkmark & \xmark\\
    FLUX-dev        & \xmark & \xmark & \xmark & \xmark & \checkmark & \xmark\\
    Jimeng AI 2.1   & \xmark & \xmark & \checkmark & \checkmark & \checkmark & \xmark\\
    \midrule
    \textbf{EasyText} & \checkmark & \checkmark & \checkmark & \checkmark & \checkmark & \checkmark\\
    \bottomrule
  \end{tabular}
  
  \caption{Functionality evaluation of EasyText in comparison to other competitors.}
  \label{tab:func_evaluation}
\end{table*}

\begin{abstract}
Generating accurate multilingual text with diffusion models has long been desired but remains challenging. Recent methods have made progress in rendering text in a single language, but rendering arbitrary languages is still an under-explored area. This paper introduces EasyText, a text rendering framework based on DiT (Diffusion Transformer), which connects denoising latents with multilingual character tokens encoded as character tokens. We propose character positioning encoding and position encoding interpolation techniques to achieve controllable and precise text rendering. Additionally, we construct a large-scale synthetic text image dataset with 1 million multilingual image-text annotations as well as a high-quality dataset of 20K annotated images, which are used for pretraining and fine-tuning respectively. Extensive experiments and evaluations demonstrate the effectiveness and advancement of our approach in multilingual text rendering, visual quality, and layout-aware text integration.
\end{abstract}

\begin{links}
    \link{Code}{https://github.com/songyiren725/EasyText}
    \link{Datasets}{https://huggingface.co/datasets/lllrrnn/EasyText}
    \link{Extended version}{https://arxiv.org/abs/2505.24417}
\end{links}

\section{Introduction}

Scene text rendering is crucial for various real-world applications. However, most existing methods such as TextDiffuser~\cite{chen2023textdiffuser,chen2024textdiffuser}, Diff-font~\cite{he2024diff} and modern commercial models like FlUX-dev~\cite{flux2024} and Ideogram~\cite{ideogram2025},  are primarily limited to English, making multilingual text rendering still a challenging task. Glyph-ByT5-V2~\cite{Glyph-byt5-v2} was one of the earliest and most representative works to enable multilingual text rendering by introducing a specially designed glyph encoder and employing a multi-stage training strategy.

Inspired by how humans learn to write, we derive several key insights: (1) Imitative writing is considerably easier than recalling—humans typically begin by mimicking before advancing to memory-based writing. (2) Once familiar with one language, humans naturally develop the ability to reproduce text in other unfamiliar languages even without understanding them—treating it more like drawing than writing. Motivated by this, we argue that training AI to “imitate” rather than “recall” is a more efficient and effective strategy for text rendering.

The task of scene text rendering faces several key challenges: (1) Multilingual character modeling is highly complex—for example, Chinese alone has over 30,000 characters. Extending to multilingual settings drastically expands the character space, making joint modeling harder, especially for rare or low-frequency characters. This is further complicated by language imbalance and font variability. (2) Text-background integration is often unnatural; existing methods struggle to blend rendered text with scene content, leading to visual artifacts such as disconnection or pasted-on effects that undermine image realism. (3) Preserving generative priors is difficult, as fine-tuning on large-scale text-image datasets improves rendering capabilities but often degrades the model’s general image generation ability.

\begin{figure*}[t]
  \centering
  \includegraphics[width=\linewidth]{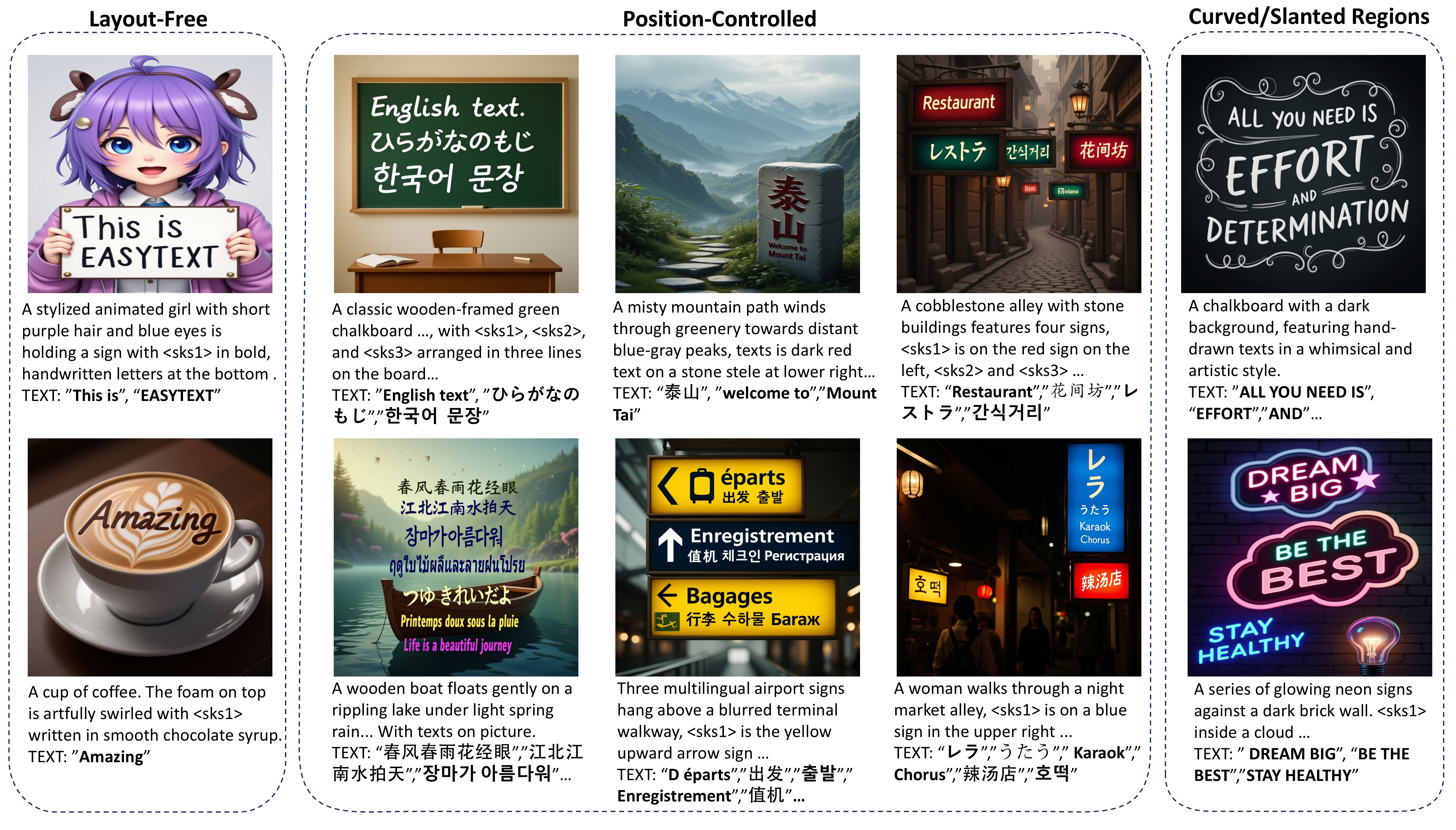}
  \captionof{figure}{Text-rendered results generated by EasyText, which supports text rendering in over ten languages and produces high-quality results. It can render text either with explicit positional control or in a layout-free manner, and effectively handles curved and slanted regions. The additional texts shown below each prompt denote the target texts to be rendered in the image which are not part of the input prompt.}
  \label{samples_general}
\end{figure*}

To this end, we present EasyText (shown in Fig. \ref{samples_general}), a multilingual text image generation framework based on Diffusion Transformers~\cite{diffusiontransformers,Attention}. We encode text into font tokens via a VAE and concatenate them in the latent space with denoised latents. Leveraging the in-context capabilities of DiT, EasyText achieves high-quality and accurate text rendering. Additionally, we propose a simple yet effective position control strategy called Implicit Character Position Alignment, which allows for precise control of character positions through positional encoding interpolation and replacement—enabling both position-aware rendering and layout-free generation.

EasyText is also highly data-efficient. Unlike Glyph-ByT5-V1/V2~\cite{Glyph-byt5,Glyph-byt5-v2}, which rely on contrastive synthetic data or massive real-world text datasets, our method simply overlays text randomly on natural images during the pretraining stage to learn glyph features. To encourage the model to learn glyph imitation rather than simple shape copying, we employ a multi-font mapping approach. Multiple different fonts are overlaid in the synthetic training images, while the condition image uses only a standard font. Afterwards, we fine-tune a lightweight LoRA~\cite{hu2022lora} on only 20K high-quality multilingual scene text images, enhancing the visual consistency between text and background. As shown in Table \ref{tab:func_evaluation}, we highlight some of the capabilities demonstrated by EasyText including: (a) \textbf{Position control}, precisely positioning text in specific locations within an image; (b) \textbf{Irregular regions}, enabling it to handle text rendering in irregular regions such as slanted or curved regions; (c) \textbf{Multilingual text handling}, supporting text generation in multiple languages such as Chinese, English, Japanese, etc; (d) \textbf{Long text rendering}, which can render extended text passages; (e) \textbf{Text-image blending}, which integrates text seamlessly into images, maintaining natural visual consistency; and (f) \textbf{Unseen character generalization}, allowing the model to exhibit generalization capability on unfamiliar and unseen characters.

In summary, our contributions are:

1. We propose EasyText, a framework that teaches AI to “imitate” rather than “recall”, achieving high-quality multilingual text rendering by harnessing the in-context learning power of Diffusion Transformers.

2. We introduce Implicit Character Position Alignment, which precisely controls text placement via position encoding operations and supports layout-free generation.

3. Extensive experiments demonstrate the effectiveness and simplicity of our method, showing superior performance on challenging scenarios such as long text, multi-text layouts, irregular regions, and unseen characters.

\section{Related Works}
\subsection{Diffusion Models}
In recent years, diffusion models~\cite{liang2024survey,song2020denoising} have achieved remarkable progress and emerged as a widely used approach for image generation, offering strong capabilities in producing high-quality and diverse visual content through iterative denoising.  Their strong capacity for modeling complex data distributions has enabled broad applicability, ranging from image synthesis and editing~\cite{brooks2023instructpix2pix,hertz2022prompt} to video generation~\cite{bar2024lumiere}. Notable examples include Stable Diffusion~\cite{rombach2022high}, its enhanced version SDXL~\cite{podell2023sdxl}, and subsequent variants, which have demonstrated the scalability and effectiveness of text-to-image diffusion models for high-quality image synthesis. Recently, the use of transformer-based denoisers—particularly the Diffusion Transformer (DiT)~\cite{diffusiontransformers, feng2023efficient, feng2025unified, feng2024evolved}, which replaces U-Net with a transformer backbone—has gained significant traction, and has been adopted in many state-of-the-art models such as FlUX-dev~\cite{flux2024}, Stable Diffusion 3.5~\cite{esser2024rectified}, and PixArt~\cite{chen2023pixart}.

\subsection{Condition-guided Diffusion Models}

Condition-guided diffusion models incorporate external signals—such as spatial structure or semantic reference—into the generative process, enabling control over layout~\cite{li2024controlnet++,zhang2025easycontrol}, content~\cite{zhang2024ssr, zhang2024fast, wang2024stablegarment, stablemakeup, stablehair, processpainter}, motion ~\cite{ma2025followfaster, ma2023glyphdraw, ma2024followyouremoji,  ma2025followyourmotion, ma2025controllable, ma2025followyourclick}, and identity~\cite{wang2024instantid}. These methods facilitate consistent synthesis, alignment with user intent, and support for flexible customization.
Early implementations based on U-Net architectures adopted two main paradigms: attention-based conditioning, which integrates semantic features via auxiliary encoders, and residual-based fusion, which injects spatial features into intermediate layers. ControlNet~\cite{zhang2023adding} and T2I-Adapter~\cite{mou2024t2i} are representative examples that improve spatial consistency and layout control under this framework.
With the adoption of transformer-based diffusion models, recent approaches reformulate both semantic and spatial conditions as token sequences and integrate them into the generation process through multi-modal attention or token concatenation mechanisms, as demonstrated in DiT-based systems like OminiControl~\cite{tan2024ominicontrol} . This unified formulation advances the development and application of conditional image generation ~\cite{wan2024grid, song2025makeanything, song2025layertracer, huang2025photodoodle, guo2025any2anytryon}, improving scalability, simplifying model design, and enabling efficient handling of multiple conditions.

\subsection{Visual Text Generation}

Text generation and rendering is a classic task, where early methods mainly relied on generative adversarial networks (GANs)~\cite{wu2019editing, cha2020few, azadi2018multi, tang2022few} and vector stroke techniques ~\cite{thamizharasan2024vecfusion, song2023clipvg, song2022clipfont}. Recent work on visual text generation with diffusion models primarily focuses on optimizing text encoding and spatial control mechanisms to improve the fidelity and controllability of rendered text. Character-aware encoders are increasingly adopted: GlyphDraw~\cite{ma2023glyphdraw}, GlyphControl~\cite{yang2023glyphcontrol}, and AnyText~\cite{tuo2023anytext} embed glyph or OCR features into conditional inputs, while TextDiffuser-2~\cite{chen2024textdiffuser} leverages character-level tokenization to improve alignment. However, semantic bias and tokenizer limitations still persist. Spatial control has been addressed by introducing explicit layout-related conditions. TextDiffuser~\cite{chen2023textdiffuser} and ControlText~\cite{jiang2025controltext} use segmentation or layout masks, and UDiffText~\cite{zhao2023udifftext} and Brush Your Text~\cite{zhang2024brush} refine attention to enforce region-level alignment. For multilingual rendering, works like AnyText~\cite{tuo2023anytext}, GlyphControl~\cite{yang2023glyphcontrol}, and Glyph-ByT5-v2~\cite{Glyph-byt5-v2} bypass tokenizers using glyph encoders or tokenizer-free models. Yet, rendering quality remains limited by encoder–diffusion compatibility. Newer models (GlyphDraw2~\cite{ma2025glyphdraw2}, JoyType~\cite{li2024joytype}, AnyText2~\cite{tuo2024anytext2}, FonTS~\cite{shi2024fonts}, RepText~\cite{wang2025reptext}) improve layout precision, while closed models (Kolors 2.0~\cite{kolors2024}, WordCon ~\cite{shi2025wordcon}, Seedream 3.0~\cite{gao2025seedream}, GPT-4o~\cite{openai2024gpt4o}) show promising results but often suffer from limited spatial controllability and suboptimal performance in multi-text layouts. Despite progress, challenges remain in complex layouts and multilingual text, motivating further integration of glyph, position, and semantic signals.

\section{Method}
\maketitle
\begin{figure*}[t]
  \centering
  \includegraphics[width=\linewidth]{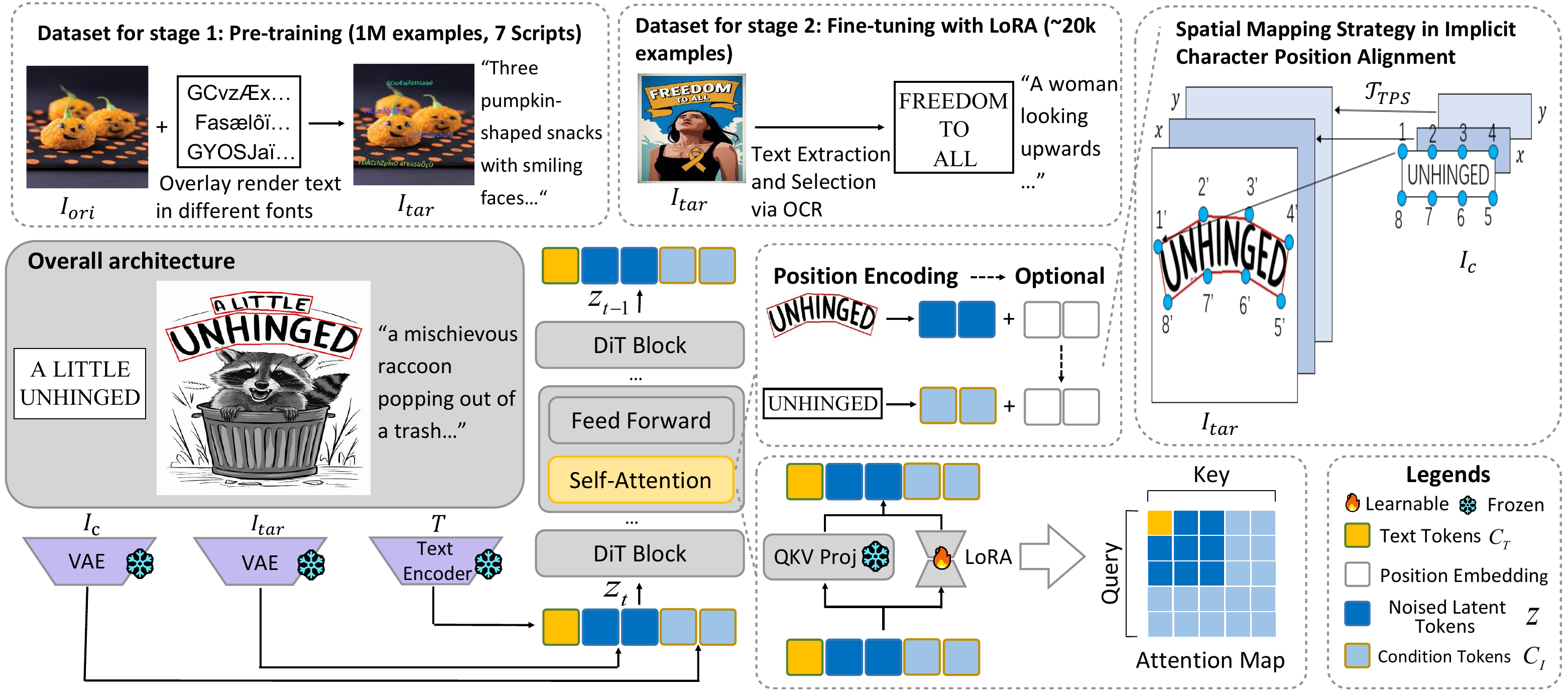}
  \caption{Overview of EasyText. A two-stage training strategy is used: large-scale pretraining for glyph generation and spatial mapping, followed by fine-tuning for visual-text integration and aesthetic refinement. Character positions from the condition input are implicitly aligned with target regions, and training proceeds with image-conditioned LoRA.}
  \label{fig:framwork}
\end{figure*}

The training of our method is based on the open-source base model FLUX (refer to Appendix Section A for details). Section~\ref{Overall Architecture} outlines the overall architecture; Section~\ref{cr} describes the target text condition representation; Section~\ref{icpa} details the Implicit Character Position Alignment; and Section~\ref{mts} explains the paired dataset construction.

\subsection{Overall Architecture}
\label{Overall Architecture}

The training pipeline adopts a two-stage strategy. The first stage involves training on a large-scale, synthetically generated dataset with balanced multilingual coverage, enabling the model to learn glyph generation across diverse scripts and accurate spatial mapping. In the second stage, we fine-tune the model on a high-quality annotated dataset to improve both the aesthetic quality of rendered text and its integration with complex scene content. For conditioning, the input condition image ($I_c$) is encoded into the same latent space as the target image ($I_{tar}$) via a VAE. Its positional encoding is aligned with the target region through Implicit Character Position Alignment, then concatenated with the target image’s position encoding before being passed to the FLUX DiT block. (Shown in Fig. \ref{fig:framwork}) Image-conditioned LoRA adaptation is then applied, with both the VAE and text encoder kept frozen.

\subsection{Character Representation in EasyFont}

\label{cr}

In contrast to conventional font conditioning approaches that typically employ symbolic or parametric representations, our methodology innovatively adopts a visually-grounded paradigm inspired by glyph morphology. The key differentiation of our character representation lies in its image-based conditioning architecture, where discrete image patches serve as fundamental conditioning units. This design philosophy stems from two critical observations:

First, we establish a unified multilingual representation to address the intrinsic typographic differences between writing systems. For alphabetic scripts (e.g., English), we use 64-pixel-high images with widths adaptive to text length, naturally capturing the connected structure of alphabetic word formations. This horizontal layout preserves crucial inter-character relationships and spacing conventions unique to Western typography. Second, for logographic systems (Chinese, Japanese, etc.), we assign a fixed-size square image to each character. For Chinese characters in particular, the image size is set to 64×64 pixels. This configuration respects the isolated nature of ideographic characters while maintaining consistent resolution. The square aspect ratio optimally captures the balanced structure inherent to characters, where stroke complexity is uniformly distributed within a square design space.

Our condition image efficiently captures the rich visual features of glyphs while providing a significantly more compact representation than many existing methods that use layout inputs matching the target image resolution. By including only the text to be rendered, it typically requires less than one-tenth of the spatial size, substantially reducing computational overhead.

\subsection{Implicit Character Position Alignment}
\label{icpa}

To enable flexible and precise spatial control over the rendered text within the condition image, we introduce an \textit{Implicit Character Position Alignment} (ICPA) mechanism, which maps the spatial coordinates of the rendered text in the target image (even in  irregular regions) onto the corresponding characters in the condition image. Given a conditional image patch $\mathbf{P}_c \in \mathbb{R}^{64 \times W_c \times 3}$ containing source typography features—where $W_c$ denotes the total width of the conditional image patch, and a target rendering region $\Omega_t$ in the output image, our method establishes position-aware feature correspondence through positional encoding extrapolation.

\textbf{Linear Alignment via Affine Transform.}
Linearly interpolate the position in the condition image to the position in the target bounding box (an axis-aligned rectangular region in this case) $\Omega_t = [x_1:x_2,y_1:y_2]$. Assuming $u \in [0,W_c-1]$ and $v \in [0,63]$ (i.e. the condition image spans indices $0$ to $W_c$ horizontally and $0$ to $63$ vertically), the affine mapping $\mathcal{T}_{\text{aff}}: (u,v)\mapsto(x,y)$ is defined as:
\begingroup
\setlength{\abovedisplayskip}{4pt}
\setlength{\belowdisplayskip}{4pt}
\begin{align}
\small
\mathcal{T}_{\text{aff}}(u,v) &= \Big( x_1 + \frac{u}{W_c - 1}(x_2 - x_1), \nonumber \\
&\quad y_1 + \frac{v}{63}(y_2 - y_1) \Big)\,,
\end{align}
\endgroup
which maps the normalized horizontal coordinate $u/(W_c-1)$ to a point between $x_1$ and $x_2$, and similarly $v/63$ to a point between $y_1$ and $y_2$, achieving a linear scaling and translation of the coordinates in condition image into the target domain. The positional encoding in condition image is then re-aligned by applying the affine transform, yielding a transformed position \((u', v') = (x, y) = \mathcal{T}_{\text{aff}}(u, v)\). These transformed coordinates are used to update the positional encoding in condition image, enabling spatial consistency between the condition and target images.

\textbf{Nonlinear Alignment via Thin-Plate Spline Interpolation.}
For nonlinear spatial alignment, we adopt Thin-Plate Spline (TPS) interpolation, which defines a smooth mapping that exactly fits a set of control point correspondences between the source patch and the target region, while minimizing second-order bending distortion. Here, \(\Omega_t\) denotes an irregular region. We obtain \(K\) control points along its boundary to form a deformation region that closely fits and fully covers the rendered text. These points act as spatial anchors for the TPS transformation. Let $\{(u_i, v_i)\}_{i=1}^K$ be $K$ landmark points in the condition image $P_c$ and $\{(x_i, y_i)\}_{i=1}^K$ their corresponding positions in the target region $\Omega_t$. The TPS mapping $\mathcal{T}_{\text{TPS}}: (u,v)\mapsto(x,y)$ can be expressed as an affine base plus a radial basis deformation term:
\begingroup
\small
\setlength{\abovedisplayskip}{4pt}
\setlength{\belowdisplayskip}{4pt}
\begin{align}
\mathcal{T}_{\text{TPS}}(u,v) 
&= A \begin{pmatrix} u \\ v \\ 1 \end{pmatrix} 
+ \sum_{i=1}^K w_i\, \phi\!\left(\sqrt{(u - u_i)^2 + (v - v_i)^2}\right), \nonumber \\
\phi(r) &= r^2 \log r,
\end{align}
\endgroup
where $A \in \mathbb{R}^{2\times 3}$ represents the affine part of the transformation and $w_i \in \mathbb{R}^2$ are the TPS warp coefficients associated with each control point. The kernel $\phi(r)=r^2\ln r$ is the fundamental solution of the biharmonic equation in 2D. The parameters $A$ and ${w_i}$ are determined by solving the interpolation constraints 
\begingroup
\setlength{\abovedisplayskip}{4pt}
\setlength{\belowdisplayskip}{4pt}
\begin{equation}
\mathcal{T}_{\text{TPS}}(u_i,v_i) = (x_i,y_i), i=1,\dots,K,
\end{equation}
\endgroup
together with additional conditions to ensure a well-posed solution ($\sum_{i=1}^K w_i = 0$, $\sum_{i=1}^K w_i u_i = 0$, $\sum_{i=1}^K w_i v_i = 0$). 
Similar to linear alignment, transformed position \((u', v') = (x, y) = \mathcal{T}_{\text{TPS}}(u, v)\) are used to update the positional encoding in condition image.

\textbf{Layout-Free Position Alignment via Positional Offset Injection.}
To enable flexible layout-free rendering, we also introduce an effective positional offset strategy. We shift the positional encoding of the conditional image by a fixed scalar offset $w_t$, which represents the width of the target image. Let $(u,v)$ index the spatial position in $\mathbf{P}_c$. Updated coordinates $(u', v')$ are obtained by applying a positional offset to the original coordinates:
\begingroup
\small
\setlength{\abovedisplayskip}{4pt}
\setlength{\belowdisplayskip}{4pt}
\begin{equation}
\begin{cases}
u' = u +  w_t \\
v' = v 
\end{cases}.
\end{equation}
\endgroup
This ensures positional encoding uniqueness without binding the conditional image to any specific target location, enabling more flexible text rendering.

\subsection{EasyText Dataset Construction}
\label{mts}

\begin{table}[t]
  \centering
  \footnotesize
  \setlength{\tabcolsep}{2.6pt}
  \begin{tabular}{lcccc}
    \toprule
    \textbf{Script Type} &
    \multicolumn{3}{c}{\textbf{Large-Scale Pre-Training}} &
    \textbf{Fine-Tuning} \\
    & Unique Chars & Font Types & Samples & Samples \\
    \midrule
    Chinese     & 7000 & 18 & 230K & 5.5K \\
    Korean      & 4308 & 13 & 120K & 0.1K \\
    Japanese    & 2922 & 17 & 120K & 0.1K \\
    Thai        & 2128 & 3  & 120K & \xmark \\
    Vietnamese  & 91   & 19 & 120K & \xmark \\
    Latin       & 104  & 30 & 180K & 15K \\
    Greek       & 79   & 19 & 120K & \xmark \\
    \bottomrule
  \end{tabular}
  \caption{Statistics of multilingual data used in large-scale pre-training and fine-tuning.}
  \label{tab:script_data_stats}
\end{table}

We construct two datasets tailored to the specific needs of the pretraining and fine-tuning stages.

\textbf{Large-Scale Synthetic Dataset.} 
This dataset is designed to provide broad coverage across scripts and languages. Target images are generated by rendering multilingual text onto background images using script-based synthesis. The dataset includes scripts such as Latin, Chinese, Korean, and others, resulting in nearly 1 million samples. To reduce redundancy and improve efficiency, languages sharing the same writing system are grouped by script—for example, English, Italian, and German are treated as a single Latin script class. The rendered text is composed of random combinations of characters from the corresponding script, covering its full character set. To encourage the model to learn generalizable glyph representations rather than simply copying shapes from the condition image, the target text is rendered using diverse fonts, while the condition image is rendered using a standardized font. This font decoupling enables robust learning of character structure across styles. Dataset statistics are summarized in Table~\ref{tab:script_data_stats}.

\textbf{High-Quality Human-Annotated Dataset.}
The second dataset consists of approximately 20k high-quality image–text box pairs, primarily in Chinese and English. Text regions are extracted using PP-OCR~\cite{du2020pp} and filtered to ensure annotation quality. In these examples, text is more naturally integrated into scene content, often with stylized typography and visually consistent stroke patterns. The rendered text is not included explicitly in the prompt, but instead triggered by placeholder tokens (e.g., \texttt{<}sks1\texttt{>}, \texttt{<}sks2\texttt{>}), supporting semantic alignment between text and background. This dataset facilitates better visual-text fusion and improves typographic aesthetics.

This two-stage training strategy reduces the need for large-scale text-image datasets, requiring only a small fine-tuning set—particularly well-suited in multilingual settings where annotated data is scarce.

\section{Experiments}

\begin{figure}[t]
  \centering
  \includegraphics[width=\linewidth]{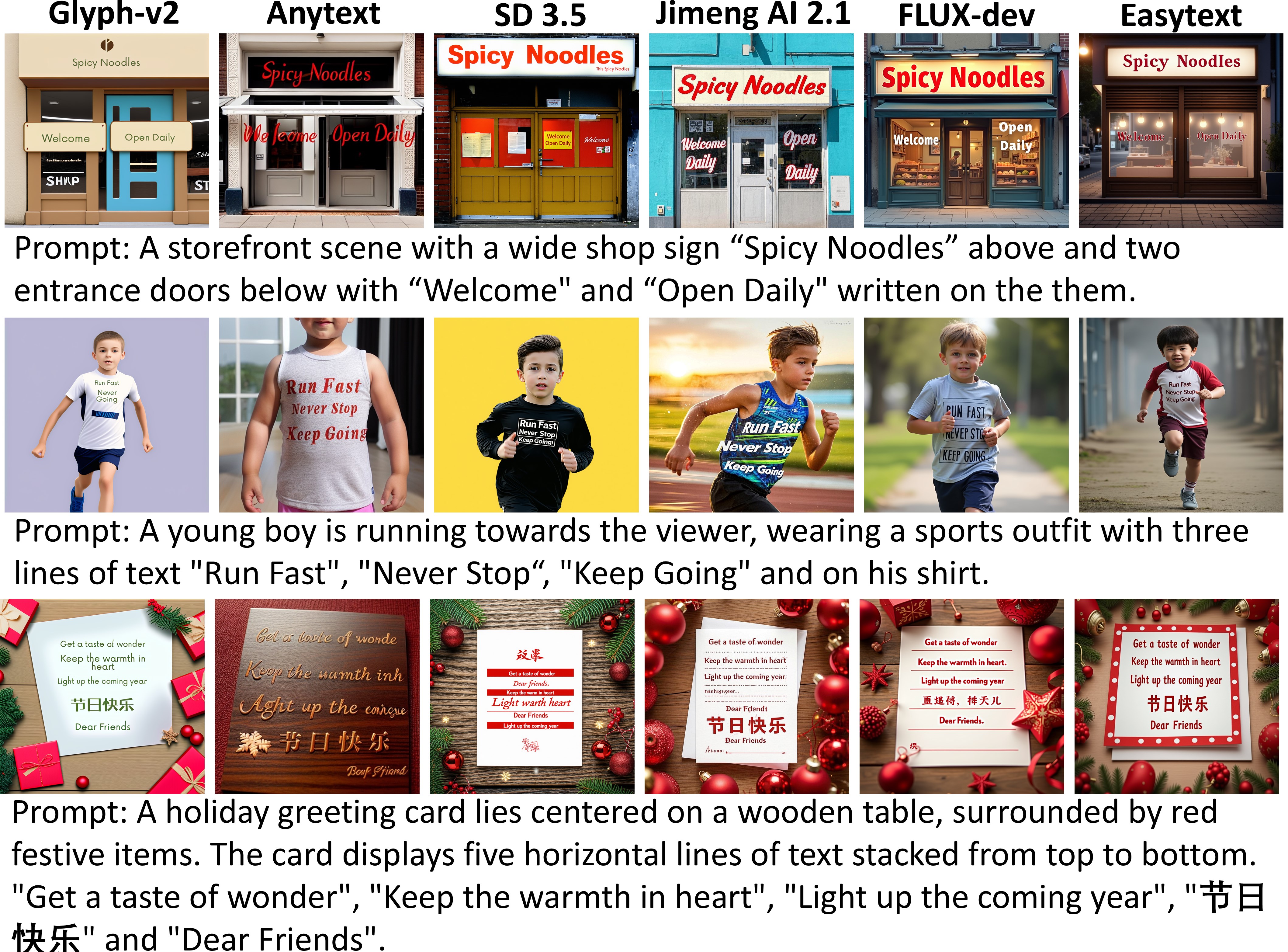}
  \caption{Qualitative comparison of EasyText with other methods, focusing on the generation quality of both text and images, reveals that EasyText demonstrates outstanding performance.}
  \label{fig:comparison}
\end{figure}

\begin{figure}[t]
  \centering
  \includegraphics[width=\linewidth]{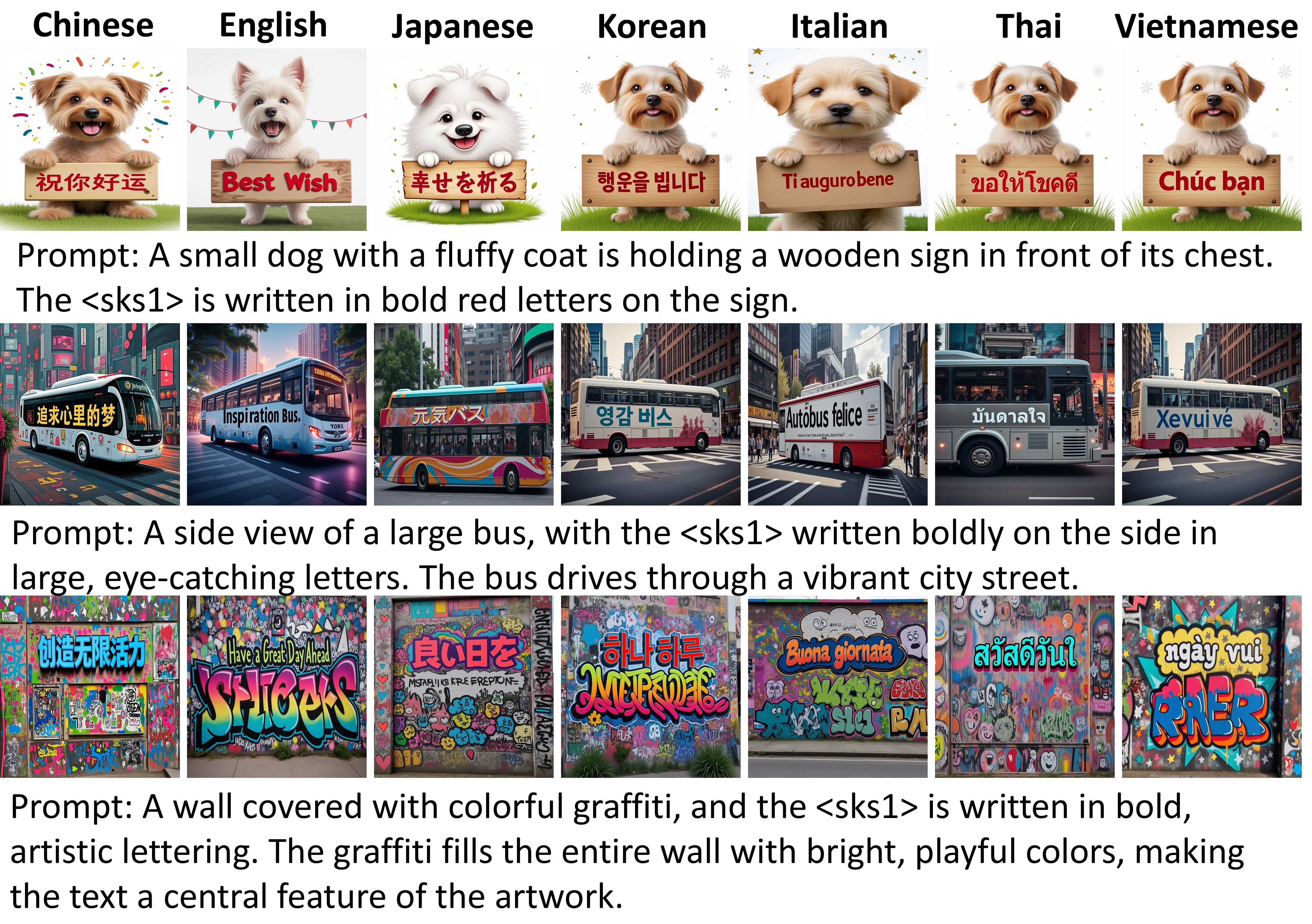}
  \caption{Qualitative comparison of EasyText across multiple languages of the same prompt.}
  \label{fig:multilingual_samples}
\end{figure}

\subsection{Experimental Settings}

\label{Experimental Settings}
\textbf{Implementation Details.} Our method is implemented based on the open-source FLUX-dev framework with LoRA-based parameter-efficient tuning. In the first stage, we set the LoRA rank to 128 and train on 4 H20 GPUs. The model is first trained at 512×512 resolution for 48,000 steps with a batch size of 24, and then at 1024×1024 for 12,000 steps. In the second stage, we reduce the LoRA rank to 32 and fine-tune at 1024 resolution for 10,000 steps with a batch size of 8. We use the AdamW optimizer with a learning rate of 1e-4.

\textbf{Evaluation Metric.} To evaluate the accuracy of multilingual text generation, we adopt two complementary metrics:
(i) Character-level precision, which measures the correctness of generated characters relative to ground-truth texts. This reflects the model's ability to produce accurate predictions at the most basic unit level for each language.
(ii) Sentence-level precision, which assesses whether the entire text line content within each controlled text box is rendered completely correctly. This captures the overall consistency and correctness at the region level.

\textbf{Multilingual Benchmark.} We construct a multilingual benchmark with 90 language-agnostic prompts. For each language, prompts are paired with language-specific text to generate condition images while preserving semantic intent. Four images are generated per prompt-language pair. For logographic scripts, the average number of characters per text box is 7.1, and for alphabetic scripts, it is 14.3. Each image contains an average of 1.7 text boxes. To evaluate the generated results, three volunteers were tasked with assessing the accuracy of the text in the images, ensuring reliable multilingual generation quality.

\subsection{Comparison Results}
\begin{table}[t]
  \centering
  \fontsize{9pt}{11pt}\selectfont
  \setlength{\tabcolsep}{4pt}
  \begin{tabular}{lcccc}
    \toprule
    \multirow{2}{*}{\textbf{Method}} &
    \multicolumn{2}{c}{\textbf{English}} &
    \multicolumn{2}{c}{\textbf{Chinese}} \\
    & Chars Pre & Sen Pre & Chars Pre & Sen Pre \\
    \midrule
    EasyText (pretrain)   & \textbf{0.9968} & \textbf{0.9813} & \textbf{0.9344} & 0.6562 \\
    EasyText (fine-tune)  & 0.9945 & 0.9625 & 0.9312 & 0.6438 \\
    Glyph-SDXL-v2         & 0.9959 & 0.9688 & 0.9258 & 0.6250 \\
    FLUX                  & 0.9180 & 0.7062 & \xmark & \xmark \\
    Stable Diffusion 3.5  & 0.9556 & 0.7938 & \xmark & \xmark \\
    Jimeng 2.1            & 0.9812 & 0.8687 & 0.9214 & \textbf{0.6813} \\
    ANYTEXT               & 0.8978 & 0.6364 & 0.8824 & 0.6071 \\
    \bottomrule
  \end{tabular}
  \caption{Character/Sentence level precision for English and Chinese across different methods.}
  \label{tab:char_line_precision}
\end{table}
We evaluate our model on a multilingual benchmark against state-of-the-art commercial models (e.g., FLUX~\cite{flux2024}, Jimeng AI~\cite{jimengai2024}, SD3.5~\cite{esser2024rectified}) and other methods (e.g., Glyph-ByT5-v2~\cite{Glyph-byt5-v2}, AnyText~\cite{tuo2023anytext}), following the benchmark and metrics outlined in Sec. \ref{Experimental Settings}. For languages with limited support in prior rendering methods, we compare primarily with Glyph-SDXL-v2. All baseline models are evaluated using their official inference settings. For each sample, four images are generated with identical image descriptions and target texts across all methods. For models lacking conditional input support, the target text is included in the prompt;  evaluation focuses solely on text accuracy, regardless of spatial layout or surrounding content.

\textbf{Comprehensive Quality Assessment.} Beyond precision evaluation, we assess generation quality using CLIPScore and OCR accuracy for objective comparison against existing baselines. We further incorporate GPT-4o assessment and a user study across four subjective criteria: Image Aesthetics, Text Aesthetics, Text Quality, and Text-Image Fusing, to evaluate overall fidelity and alignment.

The results show that our model surpasses competing methods in text rendering precision across several languages, including English and Italian (Table~\ref{tab:lang_wise_precision}) and achieves strong character-level accuracy in Chinese (Table~\ref{tab:char_line_precision}), though slightly behind Jimeng AI at the sentence level. It also performs well in unsupported languages like Thai and Greek. In Table~\ref{tab:rendering_metrics}, it also leads in OCR accuracy and improves over FLUX in CLIPScore, indicating higher visual-text alignment. These trends are further supported by GPT-4o evaluation results. After pretraining, the model demonstrates strong text rendering performance from condition images. However, it shows limited text-image coherence, as indicated by lower CLIPScore and GPT-4o evaluations. Fine-tuning alleviates this issue, significantly improving CLIPScore, Text Aesthetics, and overall visual-text alignment. 

\begin{table}[t]
  \centering
  \fontsize{9pt}{11pt}\selectfont
  \setlength{\tabcolsep}{2pt}
  \begin{tabular}{lcccccc}
    \toprule
    \multirow{2}{*}{\textbf{Language}} &
    \multicolumn{2}{c}{\textbf{EasyText (pre)}} &
    \multicolumn{2}{c}{\textbf{EasyText (fine)}} &
    \multicolumn{2}{c}{\textbf{Glyph-ByT5-v2}} \\
    & Chars & Sent & Chars & Sent & Chars & Sent \\
    \midrule
    French     & \textbf{0.9867} & 0.8552 & 0.9673 & 0.7500 & 0.9812 & \textbf{0.8625} \\
    German     & 0.9818 & \textbf{0.8333} & 0.9732 & 0.7619 & \textbf{0.9828} & 0.7976 \\
    Korean     & 0.9341 & 0.7125 & 0.9203 & 0.6713 & \textbf{0.9441} & \textbf{0.7437} \\
    Japanese   & \textbf{0.9265} & \textbf{0.7179} & 0.9194 & 0.6562 & 0.9059 & 0.6187 \\
    Italian    & \textbf{0.9740} & \textbf{0.9125} & 0.9638 & 0.8571 & 0.9385 & 0.8333 \\
    Thai       & \textbf{0.9628} & \textbf{0.7443} & 0.9334 & 0.6500 & \xmark & \xmark \\
    Vietnamese & \textbf{0.9605} & \textbf{0.7312} & 0.9401 & 0.6474 & \xmark & \xmark \\
    Greek      & \textbf{0.9702} & \textbf{0.7685} & 0.9360 & 0.6875 & \xmark & \xmark \\
    \bottomrule
  \end{tabular}
  \caption{Multilingual text generation precision with our unified and generalized model, compared to Glyph-SDXL-v2.}
  \label{tab:lang_wise_precision}
\end{table}

\subsection{Qualitative Results}
We provide qualitative comparisons of multilingual text rendering, including Chinese, English, and other languages. Results are shown in Fig. \ref{fig:multilingual_samples}. Compared to previous region-controlled text rendering methods, our approach demonstrates significant improvements in visual quality and text fidelity, with better visual-text integration, higher OCR accuracy, and enhanced aesthetic coherence (shown in Fig. \ref{fig:comparison}). Additionally, unlike commercial text-to-image models, our method allows for more precise spatial control over rendered text, particularly in generating multiple paragraphs of relatively long text (e.g., 4–5 paragraphs with around 20 characters each) while maintaining consistent layouts.
We further evaluate the model’s ability to handle challenging text generation scenarios that are typically difficult for existing methods. Specifically, our approach generalizes well to unseen characters, slanted or non-rectilinear text regions, and long text sequences, while maintaining structural consistency and legibility. (refer to Appendix Section C for details)
\begin{table}[t]
  \centering
  \small
  \setlength{\tabcolsep}{1pt}
  \renewcommand{\arraystretch}{0.9}
  \resizebox{\columnwidth}{!}{
  \begin{tabular}{lccccc}
    \toprule
    \textbf{Metric} & \textbf{Ours (pre)} & \textbf{Ours (ft)} & \textbf{Glyph-v2} & \textbf{SD3.5} & \textbf{FLUX} \\
    \midrule
    CLIPScore         & 0.3318 & 0.3486 & 0.3140 & \textbf{0.3519} & 0.3348 \\
    OCR Acc. (\%)     & 84.32  & \textbf{88.72}  & 82.33  & 79.33  & 76.09  \\
    \midrule
    \multicolumn{2}{l}{\textbf{Evaluation by GPT-4o}} & & & & \\
    Image Aesthetics     & 81.58  & 83.86  & 75.86  & \textbf{84.58}  & 83.61  \\
    Text Aesthetics      & 65.14  & \textbf{73.79}  & 65.28  & 72.06  & 71.90  \\
    Text Quality         & 84.58  & \textbf{90.66}  & 86.25  & 85.79  & 86.95  \\
    Text-Image Fusing    & 74.48  & \textbf{81.28}  & 74.80  & 81.12  & 78.66  \\
    \bottomrule
  \end{tabular}}
  \caption{Quantitative evaluation of text-to-image rendering across diverse metrics and GPT-4o-based assessments. \textbf{Ours (pre)} and \textbf{Ours (ft)} denote the pretrained and finetuned versions of our model, respectively.}
  \label{tab:rendering_metrics}
\end{table}

\subsection{Ablation Studies}

\begin{figure}[h]
  \centering
  \includegraphics[width=\linewidth]{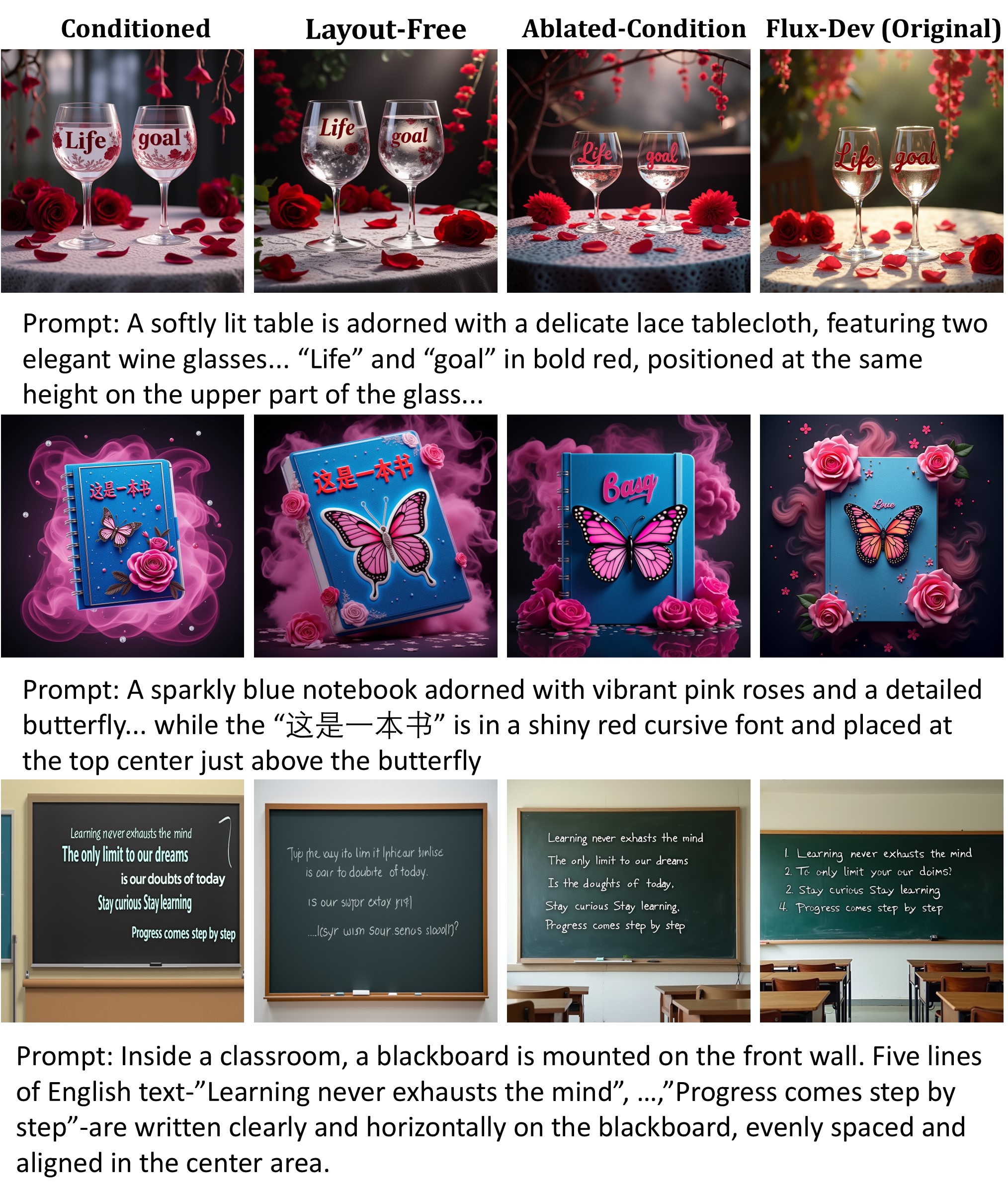}
  \caption{Ablation study comparing the full conditioned method with: (1) layout-free 
 (2) without condition inputs, and (3) original FLUX-Dev. \texttt{<sks1>}, \texttt{<sks2>} triggers are used to replace rendered text in the first two cases.}
  \label{fig:condition_abalate}
\end{figure}

To validate the effectiveness of using diverse fonts in synthetic pretraining, we conducted a controlled experiment where both the target and condition images were rendered using the same standard font. Despite high initial precision, this setup captures only direct shape mappings and fails to generalize after real-world fine-tuning. (refer to Appendix Section E) This highlights the importance of font diversity—rendering targets with multiple fonts while keeping the condition font fixed—for learning robust, transferable representations.
We compare our approach with three ablated variants: (1) removing position alignment, (2) removing the condition input, and (3) using the original FLUX model. As shown in Fig. \ref{fig:condition_abalate}, incorporating position mapping substantially improves spatial precision and text accuracy, particularly in multi-text scenarios. In comparison, when removing the condition input, the target rendered text is instead provided in the prompt(in our full method, rendered text content is given via the condition input). This demonstrates that our fine-tuned model effectively retains the strong generation capability of the base FLUX model.

\section{Limitation}

The Implicit Character Position Alignment mechanism is less effective when character positions are substantially overlapping, occasionally leading to reduced rendering accuracy. In addition, training across multiple scripts results in confusion between simple but visually similar characters from different writing systems. These cases are infrequent but consistently observed.

\section{Conclusion}

In this work, we propose EasyText, a diffusion-based framework for multilingual and controllable text generation in text-to-image synthesis. EasyText learns to mimic glyph features and supports precise and flexible text placement through implicit character alignment. With a two-stage training strategy, the method significantly reduces reliance on real multilingual data while maintaining high rendering accuracy. It also demonstrates strong visual-text integration, effectively embedding text into complex scenes. 

{\small
\bibliography{references}

@article{ma2025followfaster,
  title={Follow-your-emoji-faster: Towards efficient, fine-controllable, and expressive freestyle portrait animation},
  author={Ma, Yue and Yan, Zexuan and Liu, Hongyu and Wang, Hongfa and Pan, Heng and He, Yingqing and Yuan, Junkun and Zeng, Ailing and Cai, Chengfei and Shum, Heung-Yeung and others},
  journal={arXiv preprint arXiv:2509.16630},
  year={2025}
}

@article{ma2025controllable,
  title={Controllable Video Generation: A Survey},
  author={Ma, Yue and Feng, Kunyu and Hu, Zhongyuan and Wang, Xinyu and Wang, Yucheng and Zheng, Mingzhe and He, Xuanhua and Zhu, Chenyang and Liu, Hongyu and He, Yingqing and others},
  journal={arXiv preprint arXiv:2507.16869},
  year={2025}
}

@article{ma2025followyourmotion,
  title={Follow-Your-Motion: Video Motion Transfer via Efficient Spatial-Temporal Decoupled Finetuning},
  author={Ma, Yue and Liu, Yulong and Zhu, Qiyuan and Yang, Ayden and Feng, Kunyu and Zhang, Xinhua and Li, Zhifeng and Han, Sirui and Qi, Chenyang and Chen, Qifeng},
  journal={arXiv preprint arXiv:2506.05207},
  year={2025}
}

@inproceedings{ma2024followyouremoji,
  title={Follow-your-emoji: Fine-controllable and expressive freestyle portrait animation},
  author={Ma, Yue and Liu, Hongyu and Wang, Hongfa and Pan, Heng and He, Yingqing and Yuan, Junkun and Zeng, Ailing and Cai, Chengfei and Shum, Heung-Yeung and Liu, Wei and others},
  booktitle={SIGGRAPH Asia 2024 Conference Papers},
  pages={1--12},
  year={2024}
}

@inproceedings{ma2025followyourclick,
  title={Follow-Your-Click: Open-domain Regional Image Animation via Motion Prompts},
  author={Ma, Yue and He, Yingqing and Wang, Hongfa and Wang, Andong and Shen, Leqi and Qi, Chenyang and Ying, Jixuan and Cai, Chengfei and Li, Zhifeng and Shum, Heung-Yeung and others},
  booktitle={Proceedings of the AAAI Conference on Artificial Intelligence},
  volume={39},
  number={6},
  pages={6018--6026},
  year={2025}
}

@article{Glyph-byt5-v2,
  title={Glyph-byt5-v2: A strong aesthetic baseline for accurate multilingual visual text rendering},
  author={Liu, Zeyu and Liang, Weicong and Zhao, Yiming and Chen, Bohan and Liang, Lin and Wang, Lijuan and Li, Ji and Yuan, Yuhui},
  journal={arXiv preprint arXiv:2406.10208},
  year={2024}
}

@inproceedings{Glyph-byt5,
  title={Glyph-byt5: A customized text encoder for accurate visual text rendering},
  author={Liu, Zeyu and Liang, Weicong and Liang, Zhanhao and Luo, Chong and Li, Ji and Huang, Gao and Yuan, Yuhui},
  booktitle={European Conference on Computer Vision},
  pages={361--377},
  year={2024},
  organization={Springer}
}

@article{Attention,
  title={Attention is all you need},
  author={Vaswani, Ashish and Shazeer, Noam and Parmar, Niki and Uszkoreit, Jakob and Jones, Llion and Gomez, Aidan N and Kaiser, {\L}ukasz and Polosukhin, Illia},
  journal={Advances in neural information processing systems},
  volume={30},
  year={2017}
}

@inproceedings{diffusiontransformers,
  title={Scalable diffusion models with transformers},
  author={Peebles, William and Xie, Saining},
  booktitle={Proceedings of the IEEE/CVF international conference on computer vision},
  pages={4195--4205},
  year={2023}
}

@article{hu2022lora,
  title={Lora: Low-rank adaptation of large language models.},
  author={Hu, Edward J and Shen, Yelong and Wallis, Phillip and Allen-Zhu, Zeyuan and Li, Yuanzhi and Wang, Shean and Wang, Lu and Chen, Weizhu and others},
  journal={ICLR},
  volume={1},
  number={2},
  pages={3},
  year={2022}
}

@article{chen2023textdiffuser,
  title={Textdiffuser: Diffusion models as text painters},
  author={Chen, Jingye and Huang, Yupan and Lv, Tengchao and Cui, Lei and Chen, Qifeng and Wei, Furu},
  journal={Advances in Neural Information Processing Systems},
  volume={36},
  pages={9353--9387},
  year={2023}
}

@inproceedings{chen2024textdiffuser,
  title={Textdiffuser-2: Unleashing the power of language models for text rendering},
  author={Chen, Jingye and Huang, Yupan and Lv, Tengchao and Cui, Lei and Chen, Qifeng and Wei, Furu},
  booktitle={European Conference on Computer Vision},
  pages={386--402},
  year={2024},
  organization={Springer}
}

@article{he2024diff,
  title={Diff-font: Diffusion model for robust one-shot font generation},
  author={He, Haibin and Chen, Xinyuan and Wang, Chaoyue and Liu, Juhua and Du, Bo and Tao, Dacheng and Yu, Qiao},
  journal={International Journal of Computer Vision},
  volume={132},
  number={11},
  pages={5372--5386},
  year={2024},
  publisher={Springer}
}

@misc{flux2024,
  author       = {Black Forest Labs},
  title        = {Flux},
  year         = {2024},
  howpublished = {\url{https://github.com/black-forest-labs/flux}},
  note         = {Accessed July 10, 2025}
}

@inproceedings{esser2024rectified,
  author    = {Esser, Patrick and Kulal, S and Blattmann, Andreas and ...},
  title     = {Scaling Rectified Flow Transformers for High-Resolution Image Synthesis},
  booktitle = {Forty-first International Conference on Machine Learning (ICML)},
  year      = {2024}
}

@misc{ideogram2025,
  title        = {Ideogram v3.0},
  author       = {{Ideogram Inc.}},
  year         = {2025},
  howpublished = {\url{https://ideogram.ai/}},
  note         = {Accessed July 10, 2025}
}

@article{song2020denoising,
  title={Denoising diffusion implicit models},
  author={Song, Jiaming and Meng, Chenlin and Ermon, Stefano},
  journal={arXiv preprint arXiv:2010.02502},
  year={2020}
}

@inproceedings{liang2024survey,
  title={A Survey of Multimodel Large Language Models},
  author={Liang, Zijing and Xu, Yanjie and Hong, Yifan and Shang, Penghui and Wang, Qi and Fu, Qiang and Liu, Ke},
  booktitle={Proceedings of the 3rd International Conference on Computer, Artificial Intelligence and Control Engineering},
  pages={405--409},
  year={2024}
}

@inproceedings{brooks2023instructpix2pix,
  title={Instructpix2pix: Learning to follow image editing instructions},
  author={Brooks, Tim and Holynski, Aleksander and Efros, Alexei A},
  booktitle={Proceedings of the IEEE/CVF conference on computer vision and pattern recognition},
  pages={18392--18402},
  year={2023}
}

@article{hertz2022prompt,
  title={Prompt-to-prompt image editing with cross attention control},
  author={Hertz, Amir and Mokady, Ron and Tenenbaum, Jay and Aberman, Kfir and Pritch, Yael and Cohen-Or, Daniel},
  journal={arXiv preprint arXiv:2208.01626},
  year={2022}
}

@inproceedings{bar2024lumiere,
  title={Lumiere: A space-time diffusion model for video generation},
  author={Bar-Tal, Omer and Chefer, Hila and Tov, Omer and Herrmann, Charles and Paiss, Roni and Zada, Shiran and Ephrat, Ariel and Hur, Junhwa and Liu, Guanghui and Raj, Amit and others},
  booktitle={SIGGRAPH Asia 2024 Conference Papers},
  pages={1--11},
  year={2024}
}

@inproceedings{rombach2022high,
  title={High-resolution image synthesis with latent diffusion models},
  author={Rombach, Robin and Blattmann, Andreas and Lorenz, Dominik and Esser, Patrick and Ommer, Bj{\"o}rn},
  booktitle={Proceedings of the IEEE/CVF conference on computer vision and pattern recognition},
  pages={10684--10695},
  year={2022}
}

@article{podell2023sdxl,
  title={Sdxl: Improving latent diffusion models for high-resolution image synthesis},
  author={Podell, Dustin and English, Zion and Lacey, Kyle and Blattmann, Andreas and Dockhorn, Tim and M{\"u}ller, Jonas and Penna, Joe and Rombach, Robin},
  journal={arXiv preprint arXiv:2307.01952},
  year={2023}
}

@article{chen2023pixart,
  title={Pixart: Fast training of diffusion transformer for photorealistic text-to-image synthesis},
  author={Chen, Junsong and Yu, Jincheng and Ge, Chongjian and Yao, Lewei and Xie, Enze and Wu, Yue and Wang, Zhongdao and Kwok, James and Luo, Ping and Lu, Huchuan and others},
  journal={arXiv preprint arXiv:2310.00426},
  year={2023}
}

@inproceedings{zhang2023adding,
  title={Adding conditional control to text-to-image diffusion models},
  author={Zhang, Lvmin and Rao, Anyi and Agrawala, Maneesh},
  booktitle={Proceedings of the IEEE/CVF international conference on computer vision},
  pages={3836--3847},
  year={2023}
}

@inproceedings{mou2024t2i,
  title={T2i-adapter: Learning adapters to dig out more controllable ability for text-to-image diffusion models},
  author={Mou, Chong and Wang, Xintao and Xie, Liangbin and Wu, Yanze and Zhang, Jian and Qi, Zhongang and Shan, Ying},
  booktitle={Proceedings of the AAAI conference on artificial intelligence},
  volume={38},
  number={5},
  pages={4296--4304},
  year={2024}
}

@article{ma2023glyphdraw,
  title={Glyphdraw: Seamlessly rendering text with intricate spatial structures in text-to-image generation},
  author={Ma, Jian and Zhao, Mingjun and Chen, Chen and Wang, Ruichen and Niu, Di and Lu, Haonan and Lin, Xiaodong},
  journal={arXiv preprint arXiv:2303.17870},
  year={2023}
}

@article{yang2023glyphcontrol,
  title={Glyphcontrol: glyph conditional control for visual text generation},
  author={Yang, Yukang and Gui, Dongnan and Yuan, Yuhui and Liang, Weicong and Ding, Haisong and Hu, Han and Chen, Kai},
  journal={Advances in Neural Information Processing Systems},
  volume={36},
  pages={44050--44066},
  year={2023}
}

@article{tuo2023anytext,
  title={Anytext: Multilingual visual text generation and editing},
  author={Tuo, Yuxiang and Xiang, Wangmeng and He, Jun-Yan and Geng, Yifeng and Xie, Xuansong},
  journal={arXiv preprint arXiv:2311.03054},
  year={2023}
}

@article{jiang2025controltext,
  title={ControlText: Unlocking Controllable Fonts in Multilingual Text Rendering without Font Annotations},
  author={Jiang, Bowen and Yuan, Yuan and Bai, Xinyi and Hao, Zhuoqun and Yin, Alyson and Hu, Yaojie and Liao, Wenyu and Ungar, Lyle and Taylor, Camillo J},
  journal={arXiv preprint arXiv:2502.10999},
  year={2025}
}

@article{zhao2023udifftext,
  title={Udifftext: A unified framework for high-quality text synthesis in arbitrary images via character-aware diffusion models},
  author={Zhao, Yiming and Lian, Zhouhui},
  journal={arXiv preprint arXiv:2312.04884},
  year={2023}
}

@inproceedings{zhang2024brush,
  title={Brush your text: Synthesize any scene text on images via diffusion model},
  author={Zhang, Lingjun and Chen, Xinyuan and Wang, Yaohui and Lu, Yue and Qiao, Yu},
  booktitle={Proceedings of the AAAI Conference on Artificial Intelligence},
  volume={38},
  number={7},
  pages={7215--7223},
  year={2024}
}

@inproceedings{ma2025glyphdraw2,
  title={Glyphdraw2: Automatic generation of complex glyph posters with diffusion models and large language models},
  author={Ma, Jian and Deng, Yonglin and Chen, Chen and Du, Nanyang and Lu, Haonan and Yang, Zhenyu},
  booktitle={Proceedings of the AAAI Conference on Artificial Intelligence},
  volume={39},
  number={6},
  pages={5955--5963},
  year={2025}
}

@article{li2024joytype,
  title={JoyType: A Robust Design for Multilingual Visual Text Creation},
  author={Li, Chao and Jiang, Chen and Liu, Xiaolong and Zhao, Jun and Wang, Guoxin},
  journal={arXiv preprint arXiv:2409.17524},
  year={2024}
}

@article{tuo2024anytext2,
  title={AnyText2: Visual Text Generation and Editing With Customizable Attributes},
  author={Tuo, Yuxiang and Geng, Yifeng and Bo, Liefeng},
  journal={arXiv preprint arXiv:2411.15245},
  year={2024}
}

@misc{openai2024gpt4o,
  author       = {{OpenAI}},
  title        = {{GPT-4o}},
  year         = {2024},
  howpublished = {\url{https://chatgpt.com/}},
  note         = {Accessed July 10, 2025}
}

@misc{kolors2024,
  author       = {{Kolors Team}},
  title        = {{Kolors 2.0}},
  year         = {2024},
  howpublished = {\url{https://github.com/Kwai-Kolors/Kolors}},
  note         = {Accessed July 10, 2025}
}

@article{gao2025seedream,
  title={Seedream 3.0 Technical Report},
  author={Gao, Yu and Gong, Lixue and Guo, Qiushan and Hou, Xiaoxia and Lai, Zhichao and Li, Fanshi and Li, Liang and Lian, Xiaochen and Liao, Chao and Liu, Liyang and others},
  journal={arXiv preprint arXiv:2504.11346},
  year={2025}
}

@article{tan2024ominicontrol,
  title={Ominicontrol: Minimal and universal control for diffusion transformer},
  author={Tan, Zhenxiong and Liu, Songhua and Yang, Xingyi and Xue, Qiaochu and Wang, Xinchao},
  journal={arXiv preprint arXiv:2411.15098},
  year={2024}
}

@article{feng2024evolved,
  title={Evolved Hierarchical Masking for Self-Supervised Learning},
  author={Feng, Zhanzhou and Zhang, Shiliang},
  journal={IEEE Transactions on Pattern Analysis and Machine Intelligence},
  year={2024},
  publisher={IEEE}
}

@article{feng2023efficient,
  title={Efficient vision transformer via token merger},
  author={Feng, Zhanzhou and Zhang, Shiliang},
  journal={IEEE Transactions on Image Processing},
  volume={32},
  pages={4156--4169},
  year={2023},
  publisher={IEEE}
}

@inproceedings{feng2025unified,
  title={Unified Video Generation via Next-Set Prediction in Continuous Domain},
  author={Feng, Zhanzhou and Guo, Qingpei and Xiao, Xinyu and Xu, Ruihan and Yang, Ming and Zhang, Shiliang},
  booktitle={Proceedings of the IEEE/CVF International Conference on Computer Vision},
  pages={19427--19438},
  year={2025}
}

@article{du2020pp,
  title={Pp-ocr: A practical ultra lightweight ocr system},
  author={Du, Yuning and Li, Chenxia and Guo, Ruoyu and Yin, Xiaoting and Liu, Weiwei and Zhou, Jun and Bai, Yifan and Yu, Zilin and Yang, Yehua and Dang, Qingqing and others},
  journal={arXiv preprint arXiv:2009.09941},
  year={2020}
}

@misc{jimengai2024,
  title        = {Jimeng AI 2.1},
  author       = {{ByteDance}},
  year         = {2024},
  howpublished = {\url{https://jimeng.jianying.com/}},
  note         = {Accessed July 10, 2025}
}

@article{shi2025wordcon,
  title={WordCon: Word-level Typography Control in Scene Text Rendering},
  author={Shi, Wenda and Song, Yiren and Rao, Zihan and Zhang, Dengming and Liu, Jiaming and Zou, Xingxing},
  journal={arXiv preprint arXiv:2506.21276},
  year={2025}
}

@inproceedings{li2024controlnet++,
  title={ControlNet++: Improving Conditional Controls with Efficient Consistency Feedback: Project Page: liming-ai. github. io/ControlNet\_Plus\_Plus},
  author={Li, Ming and Yang, Taojiannan and Kuang, Huafeng and Wu, Jie and Wang, Zhaoning and Xiao, Xuefeng and Chen, Chen},
  booktitle={European Conference on Computer Vision},
  pages={129--147},
  year={2024},
  organization={Springer}
}

@article{wang2025reptext,
  title={RepText: Rendering Visual Text via Replicating},
  author={Wang, Haofan and Xu, Yujia and Li, Yimeng and Li, Junchen and Zhang, Chaowei and Wang, Jing and Yang, Kejia and Chen, Zhibo},
  journal={arXiv preprint arXiv:2504.19724},
  year={2025}
}

@inproceedings{song2022clipfont,
  title={CLIPFont: Text Guided Vector WordArt Generation.},
  author={Song, Yiren and Zhang, Yuxuan},
  booktitle={BMVC},
  pages={543},
  year={2022}
}

@article{zhang2025easycontrol,
  title={Easycontrol: Adding efficient and flexible control for diffusion transformer},
  author={Zhang, Yuxuan and Yuan, Yirui and Song, Yiren and Wang, Haofan and Liu, Jiaming},
  journal={arXiv preprint arXiv:2503.07027},
  year={2025}
}

@inproceedings{zhang2024ssr,
  title={Ssr-encoder: Encoding selective subject representation for subject-driven generation},
  author={Zhang, Yuxuan and Song, Yiren and Liu, Jiaming and Wang, Rui and Yu, Jinpeng and Tang, Hao and Li, Huaxia and Tang, Xu and Hu, Yao and Pan, Han and others},
  booktitle={Proceedings of the IEEE/CVF Conference on Computer Vision and Pattern Recognition},
  pages={8069--8078},
  year={2024}
}

@inproceedings{zhang2024fast,
  title={Fast personalized text to image synthesis with attention injection},
  author={Zhang, Yuxuan and Song, Yiren and Yu, Jinpeng and Pan, Han and Jing, Zhongliang},
  booktitle={ICASSP 2024-2024 IEEE International Conference on Acoustics, Speech and Signal Processing (ICASSP)},
  pages={6195--6199},
  year={2024},
  organization={IEEE}
}

@article{song2025makeanything,
  title={MakeAnything: Harnessing Diffusion Transformers for Multi-Domain Procedural Sequence Generation},
  author={Song, Yiren and Liu, Cheng and Shou, Mike Zheng},
  journal={arXiv preprint arXiv:2502.01572},
  year={2025}
}

@article{song2025layertracer,
  title={LayerTracer: Cognitive-Aligned Layered SVG Synthesis via Diffusion Transformer},
  author={Song, Yiren and Chen, Danze and Shou, Mike Zheng},
  journal={arXiv preprint arXiv:2502.01105},
  year={2025}
}

@article{huang2025photodoodle,
  title={PhotoDoodle: Learning Artistic Image Editing from Few-Shot Pairwise Data},
  author={Huang, Shijie and Song, Yiren and Zhang, Yuxuan and Guo, Hailong and Wang, Xueyin and Shou, Mike Zheng and Liu, Jiaming},
  journal={arXiv preprint arXiv:2502.14397},
  year={2025}
}

@article{processpainter,
  title={ProcessPainter: Learn Painting Process from Sequence Data},
  author={Song, Yiren and Huang, Shijie and Yao, Chen and Ye, Xiaojun and Ci, Hai and Liu, Jiaming and Zhang, Yuxuan and Shou, Mike Zheng},
  journal={arXiv preprint arXiv:2406.06062},
  year={2024}
}

@article{stablemakeup,
  title={Stable-Makeup: When Real-World Makeup Transfer Meets Diffusion Model},
  author={Zhang, Yuxuan and Wei, Lifu and Zhang, Qing and Song, Yiren and Liu, Jiaming and Li, Huaxia and Tang, Xu and Hu, Yao and Zhao, Haibo},
  journal={arXiv preprint arXiv:2403.07764},
  year={2024}
}

@article{stablehair,
  title={Stable-Hair: Real-World Hair Transfer via Diffusion Model},
  author={Zhang, Yuxuan and Zhang, Qing and Song, Yiren and Liu, Jiaming},
  journal={arXiv preprint arXiv:2407.14078},
  year={2024}
}

@article{wan2024grid,
  title={Grid: Visual layout generation},
  author={Wan, Cong and Luo, Xiangyang and Cai, Zijian and Song, Yiren and Zhao, Yunlong and Bai, Yifan and He, Yuhang and Gong, Yihong},
  journal={arXiv preprint arXiv:2412.10718},
  year={2024}
}

@article{guo2025any2anytryon,
  title={Any2AnyTryon: Leveraging Adaptive Position Embeddings for Versatile Virtual Clothing Tasks},
  author={Guo, Hailong and Zeng, Bohan and Song, Yiren and Zhang, Wentao and Zhang, Chuang and Liu, Jiaming},
  journal={arXiv preprint arXiv:2501.15891},
  year={2025}
}

@article{wang2024instantid,
  title={Instantid: Zero-shot identity-preserving generation in seconds},
  author={Wang, Qixun and Bai, Xu and Wang, Haofan and Qin, Zekui and Chen, Anthony and Li, Huaxia and Tang, Xu and Hu, Yao},
  journal={arXiv preprint arXiv:2401.07519},
  year={2024}
}

@article{shi2024fonts,
  title={FonTS: Text Rendering with Typography and Style Controls},
  author={Shi, Wenda and Song, Yiren and Zhang, Dengming and Liu, Jiaming and Zou, Xingxing},
  journal={arXiv preprint arXiv:2412.00136},
  year={2024}
}

@inproceedings{song2023clipvg,
  title={Clipvg: Text-guided image manipulation using differentiable vector graphics},
  author={Song, Yiren and Shao, Xuning and Chen, Kang and Zhang, Weidong and Jing, Zhongliang and Li, Minzhe},
  booktitle={Proceedings of the AAAI conference on artificial intelligence},
  volume={37},
  number={2},
  pages={2312--2320},
  year={2023}
}

@inproceedings{thamizharasan2024vecfusion,
  title={Vecfusion: Vector font generation with diffusion},
  author={Thamizharasan, Vikas and Liu, Difan and Agarwal, Shantanu and Fisher, Matthew and Gharbi, Micha{\"e}l and Wang, Oliver and Jacobson, Alec and Kalogerakis, Evangelos},
  booktitle={Proceedings of the IEEE/CVF Conference on Computer Vision and Pattern Recognition},
  pages={7943--7952},
  year={2024}
}

@inproceedings{azadi2018multi,
  title={Multi-content gan for few-shot font style transfer},
  author={Azadi, Samaneh and Fisher, Matthew and Kim, Vladimir G and Wang, Zhaowen and Shechtman, Eli and Darrell, Trevor},
  booktitle={Proceedings of the IEEE conference on computer vision and pattern recognition},
  pages={7564--7573},
  year={2018}
}

@inproceedings{cha2020few,
  title={Few-shot compositional font generation with dual memory},
  author={Cha, Junbum and Chun, Sanghyuk and Lee, Gayoung and Lee, Bado and Kim, Seonghyeon and Lee, Hwalsuk},
  booktitle={Computer Vision--ECCV 2020: 16th European Conference, Glasgow, UK, August 23--28, 2020, Proceedings, Part XIX 16},
  pages={735--751},
  year={2020},
  organization={Springer}
}

@inproceedings{wu2019editing,
  title={Editing text in the wild},
  author={Wu, Liang and Zhang, Chengquan and Liu, Jiaming and Han, Junyu and Liu, Jingtuo and Ding, Errui and Bai, Xiang},
  booktitle={Proceedings of the 27th ACM international conference on multimedia},
  pages={1500--1508},
  year={2019}
}

@inproceedings{tang2022few,
  title={Few-shot font generation by learning fine-grained local styles},
  author={Tang, Licheng and Cai, Yiyang and Liu, Jiaming and Hong, Zhibin and Gong, Mingming and Fan, Minhu and Han, Junyu and Liu, Jingtuo and Ding, Errui and Wang, Jingdong},
  booktitle={Proceedings of the IEEE/CVF conference on computer vision and pattern recognition},
  pages={7895--7904},
  year={2022}
}

@article{wang2024stablegarment,
  title={Stablegarment: Garment-centric generation via stable diffusion},
  author={Wang, Rui and Guo, Hailong and Liu, Jiaming and Li, Huaxia and Zhao, Haibo and Tang, Xu and Hu, Yao and Tang, Hao and Li, Peipei},
  journal={arXiv preprint arXiv:2403.10783},
  year={2024}
}
}

\end{document}